\pdfoutput=1

\documentclass[11pt]{article}

\usepackage[final]{coling}

\usepackage{times}
\usepackage{latexsym}

\usepackage[T1]{fontenc}

\usepackage[utf8]{inputenc}

\usepackage{microtype}

\usepackage{inconsolata}

\usepackage{pgf}
\usepackage{tcolorbox}
\usepackage{xspace}
\usepackage{courier}
\usepackage{pgfplots}
\usepackage{graphicx}
\usepackage{tabularx}
\usepackage{hyperref}
\usepackage{url}
\usepackage{booktabs}
\usepackage{listings}
\usepackage{fancyvrb}
\usepackage{fancybox}
\usepackage{xcolor}
\usepackage{amsmath}
\usepackage{fontawesome}
\usepackage{colortbl}
\usepackage{multirow}
\usepackage{cleveref}
\usepackage{hhline} 
\usepackage{highlight}
\usepackage{paralist}
\tcbuselibrary{most}
\usepackage{longtable} 
\renewcommand{\opacity}{50}

\pgfplotsset{compat=1.18}

\newtcolorbox{prompt}{
  enhanced,
  colback=gray!10,
  colframe=gray!50,
  arc=0mm,
  boxrule=0.5pt,
  left=5mm,
  right=5mm,
  top=2mm,
  bottom=2mm,
  fonttitle=\bfseries,
  coltitle=black,
  breakable,
  fontupper=\footnotesize\ttfamily,
      overlay={%
        \ifcase\tcbsegmentstate
        \or%
        \else%
        \fi%
    }

}

\newcommand{\verbcell}[1]{%
  \begin{Verbatim}[commandchars=\\\{\},breaklines=true,breakanywhere=true]
 #1
  \end{Verbatim}%
}

\newcommand{\ex}[1]{\textit{#1}\xspace}

\title{Learning Robust Negation Text Representations}

\author{
 \textbf{Thinh Hung Truong\textsuperscript{1}},
 \textbf{Karin Verspoor\textsuperscript{2,1}},
 \textbf{Trevor Cohn\textsuperscript{1,\thanks{~Also at Google Research.}}}
 \textbf{Timothy Baldwin\textsuperscript{3,1}} 
\\
 \textsuperscript{1}The University of Melbourne,
 \textsuperscript{2}RMIT University,
 \textsuperscript{3}MBZUAI
\\
    \href{mailto:hungthinht@student.unimelb.edu.au}{hungthinht@student.unimelb.edu.au}
}

\begin{document}
\maketitle
\begin{abstract}
Despite rapid adoption of autoregressive large language models, smaller text encoders still play an important role in text understanding tasks that require rich contextualized representations.
Negation is an important semantic function that is still not properly captured by such methods, affecting many downstream applications relying on text embeddings.
We propose a strategy to improve negation robustness of text encoders, by distilling data from large language models using diverse patterns of negation and hedging.
We adopt a standard contrastive learning strategy to finetune a strong BERT-based model, and observe large improvement in negation understanding capabilities while maintaining competitive performance on general benchmarks. 
In addition, we also show that our method can be adapted to LLMs, leading to improved performance on negation benchmarks.
\end{abstract}

\section{Introduction}

Modeling negation is an ongoing problem that text encoders still struggle with. 
For instance, embedding vectors of minimal negation pairs (\ex{I go to school} vs.\ \ex{I do not go to school}) have high cosine similarity \citep{ettinger-2020-bert, anschutz-etal-2023-correct}, despite their contradictory meaning.
This is due to the ``distributional hypothesis'' \citep{harris1954distributional} underlying text embedding methods, which learn the representation of words based on surrounding context. While highly effective in general, resulting models are insensitive to negation and related phenomena such as antonymy \citep{mrkvsic2016counter}.
This can lead to semantic anomalies in downstream applications, 
e.g.\ 
when searching for products lacking certain properties \citep{merra2023improving} and for exclusion-type queries \citep{zhang2024excluirexclusionaryneuralinformation}.
In a broader sense, negation is closely related to hedging, used to expressed ambiguity, probability, or uncertainty rather than completely refute a premise like negation.
Hedging is an even less explored topic in embedding research but is crucial to many language understanding tasks.
For instance, hedging is ubiquitous in scientific publications \citep{CROMPTON1997271, pei-jurgens-2021-measuring}, where precise stipulation of the degree of certainty in hypotheses, findings and conclusions (e.g., \ex{clear/weak/no evidence for \ldots }) is a critical component of scientific discourse, but again is not generally captured well in embedding vectors, as demonstrated in \Cref{fig:heatmap}.

\begin{figure}[t]
\centering
\begin{tabular}{m{4cm}cc}
&   \footnotesize{\textbf{MPNet}} &  \footnotesize{\textbf{H-MPNet}} \\
\midrule
\ex{Global warming is a hoax.} & \gradient{0.81} & \gradient{0.39} \\
\hline
\ex{There is not enough evidence to claim that global warming is real.} & \gradient{0.72} & \gradient{0.58}  \\
\hline
\ex{There is no doubt that global warming is real.} & \gradient{0.78} & \gradient{0.85}  \\
\bottomrule
\end{tabular}
\caption{Cosine similarities between the sentence \textit{Global warming is real.} and topically-similar sentences conveying different levels of modality, as obtained by MPNet, a strong sentence transformer,
and HedgeMPNet (H-MPNet), our model finetuned on HedgeTriple.}
\label{fig:heatmap}
\end{figure}

Modern large language models (LLMs) are highly effective across a wide range of tasks \citep[\emph{inter alia}]{openai2024gpt4technicalreport, geminiteam2024geminifamilyhighlycapable}.
Despite this, text encoders (e.g.\ BERT-based models) are still widely used for text understanding tasks, as: (1) the autoregressive nature of LLMs makes them sub-optimal for learning rich contextual text representations (cf.\ conditioned on surrounding contexts in bidirectional encoder models); (2) for classification tasks with some amount of labeled data, smaller finetuned text encoder models tend to perform better than LLMs; and (3) text encoders are a critical component of RAG systems, where embedding vectors from text encoders are used in the text retrieval stage to enhance robustness and reduce hallucination \citep{lewis2020retrieval}.

LLMs have the ability to reliably follow instructions to generate fluent text outputs, which kickstarted a line of research on synthetic data distilled from large LLMs to improve smaller, customized models \citep{eldan2023tinystoriessmalllanguagemodels, wang-etal-2023-self-instruct}.
In this work, we explore the use of synthetic data to make text encoder models more robust to negation and hedging by further finetuning text embedding models on contrastive triples distilled from a LLM.
Our contributions in this work are:
\begin{compactitem}
\item We propose a data synthesis method that is well-grounded in the linguistics literature on negation and hedging. 
\item  We show that finetuning a text encoder model on synthetic data can significantly improve its performance on negation benchmarks while preserving comparable performance on general benchmarks. Moreover, results with comparable models show the importance of data diversity over quantity.
\item  We adapt the method to decoder-only LLMs, showing improved negation understanding, with local degradation on benchmarks.


\end{compactitem}

\section{Related work}
\label{sec:related-work}

To obtain better text representations, a common and effective strategy is large-scale contrastive fine-tuning.
Specific to improving negation understanding, 
two works 
are most relevant, 
both following the method of first creating minimal pairs which differ only in negation cues, then finetuning a general-purpose text encoder to better differentiate between these pairs.
\citet{anschutz-etal-2023-correct} employed a rule-based negator to add verbal negation, modifying the main clause of the sentences by leveraging part-of-speech information (or removing negation cues if they were found in the original).
Instruction-tuned LLMs 
have also facilitated large-scale generation of synthetic data.
\citet{günther2023jina} used GPT-3.5 to negate sentence from NLI samples, with specific direction to keep the pairs ``syntactically very similar'', 
also resulting in verbal negation minimal pairs.

Although not directly related to adding negation, \citet{rezaei2024paraphrasing} also use GPT-3.5 to paraphrase negated samples in NLP benchmarks into affirmative versions, with the motivation that models process affirmative texts better than negation.
\citet{jang2023can} explore the abilities of LLMs to follow prompts containing negation, based on manual prompt modifications from a small subset of common benchmarks.
To ensure coverage of diverse negation types, \citet{truong-etal-2022-another}  manually created a small testbed for a broad class of different negation types.
To our knowledge, our work is the first to explore a taxonomy-based approach with the aim of generating negation and hedging data at large-scale.

There has also been research on improving the negation understanding of transformer models by modifying the pre-training objective. \citet{hosseini-etal-2021-understanding} use unlikelihood training to penalize the likelihood of tokens that are false in a negated sentence.
\citet{truong-etal-2022-improving} add a new mask token to explicitly mask the negation cue in sentences to learn better representations. 

Due to the prevalance of hedging in scientific communication, it is mostly explored in the science domain as an uncertainty detection task.
The BioScope dataset \citep{vincze2008bioscope}  includes negation and hedging annotations.
\citet{hedgepeer} curate a large scale uncertainty detection dataset from open-access reviews available in the open review platform, containing the most unqiue hedge cues.
To model hedging, \citet{pei-jurgens-2021-measuring} introduce a dataset containing sentence- and aspect-level certainty in scientific findings.
The work reveals that hedge words alone are not enough to model certainty.
For instance, ``\ex{Further research is necessary to understand whether this is a causal relationship}'' contains 0 hedges but has a high level of uncertainty.
This motivates us to
employ LLMs to obtain more diverse patterns rather than a template-based approach.


Our work builds on two core ideas from previous work: (1) we use LLMs to create synthetic data, but ground the generation step with clear linguistic instructions; and (2) we adopt a simple contrastive learning strategy to finetune a strong text encoder using the generated data. Beyond achieving large improvements on negation benchmarks, we demonstrate that the strategy retains general capabilities.

\section{Method}

\subsection{HedgeTriple dataset}
To make the encoder more sensitive to negation and hedging, we adopt contrastive learning. 
The crucial part for any contrastive learning algorithm is to collect positive and negative samples. 
As detailed below, given an affirmative sentence (e.g.\ \ex{I will go to school}), we assume that a hedged variant (e.g.\ \ex{I will probably go to school}) is more similar in meaning to the original than the negated text (e.g.\ \ex{I will not go to school}).
This relationship motivates the use of contrastive learning, 
minimizing the distance between an affirmative anchor and its hedged variant, while maximizing the distance 
to
the anchor's negated variant in the latent space.

\subsubsection{Selecting anchor sentences}
We select 50K anchors from the negation triples dataset\footnote{\url{https://hf.co/datasets/jinaai/negation-dataset-v2}} which was used to train Jina Embedding \citep{günther2023jina}, a competitive BERT-based encoder model.
The anchors are sourced from five common datasets used in training embeddings --- SNLI \citep{bowman-etal-2015-large}, Multi-NLI \citep{williams-etal-2018-broad}, sentence-compression \citep{filippova-altun-2013-overcoming}, Simple Wikipedia \citep{coster-kauchak-2011-simple}, and COCO Captions \citep{coco} --- to ensure general representations across different text types.
Most sentences are affirmative and around 50\% of the anchors  have simple grammatical structures with less than 15 words while complex samples with multiple clauses (> 30 words) account for around 10\%.

\subsubsection{Synthesizing negation and hedging}
\label{sec:hedge-triple-dataset}

\paragraph{Negation}
For negation, we adopt the taxonomy of negation from \citet{pullum_huddleston_huddleston_pullum_2002}. It provides linguistic definitions along with a sample of typical constructions  corresponding to each category of negation.
We made explicit how a sentence can be modified to exhibit each type of negation, 
to compensate for lack of LLM knowledge about some negation types.
The chosen prompt, along with an example \text{\ttfamily\small\textbf{input text}} and \text{\ttfamily\small\color{brown}{LLM generated text}},\footnote{We use GPT-3.5 for all prompts in \Cref{sec:hedge-triple-dataset}.} is as follows:

\begin{prompt}
Text: \textbf{A yellow and black plane is flying in the clouds and blue sky.}\\

Negate the text. The types of negation:\\

1. "verbal": verbal negation: when the negation is grammatically associated with the verb, the head of the clause.\\
2. "absolute": Absolute negator: no (including compounds nobody, nothing, etc., and the independent form none), neither, nor, never.\\
3. "affixal": Affixal negators: un-, in-, non-, -less, etc.\\
4. "lexical": Lexical negation: when the negation is added by substituting the main predicate of the sentence with its antonym or word carrying negative meaning.\\

Modified text: \\
1. "verbal": \color{brown}{The yellow plane is not flying in the clouds and blue sky}\\
\color{brown}{2. "absolute": There is no yellow and black plane flying in the clouds and blue sky.}\\
\color{brown}{3. "affixal": The non-flying plane is not in the clouds and blue sky.}\\
\color{brown}{4. "lexical": The yellow plane is grounded.}
\end{prompt}

\paragraph{Hedging}
For hedging, we adopt the hedging taxonomy proposed by \citet{CROMPTON1997271} for scientific findings, which classifies hedges based on their part-of-speech.
We  
simplify the taxonomy 
into two categories: single-word cues (e.g.\ \ex{probably}), and multi-word cues (e.g.\ \ex{it is unclear if}).
Even so, we found  that LLMs fall back to using specific cues for each type, resulting in very limited diversity of hedging in the generated texts.
To address this, we curated a list of cues (134 single-word cues and 45 multi-word cues) from the HedgePeer dataset \citep{hedgepeer} and explicitly included a random cue in the prompt for each call to the LLM (full list in \Cref{sec:hedge_cues}).
An example prompt with input and output is as follows:

\begin{prompt}
Text: \textbf{A yellow and black plane is flying in the clouds and blue sky.}\\

Add hedging to the text. Two types of hedging:\\
1. "word": single-word cue such as \textbf{reportedly}\\
2. "phrase": multi-word cue such as \textbf{not entirely clear}\\

Modified text:\\
1. "word": \color{brown}{A yellow and black plane is reportedly flying in the clouds and blue sky.}\\
\color{brown}{2. "phrase": It's not entirely \color{brown}{clear what's happening, but a yellow and black plane appears to be flying in the clouds and blue sky.}}
\end{prompt}

\subsubsection{Constructing triples}

We perform a post-processing step to filter out all samples where the generated text is too different from the anchor text (based on Levenshtein distance, with the upper threshold of 60, equivalent to ~10 words) and retain only minimal pairs.
This is an essential step to ensure that the triples are still topically similar.
For instance, the pair \ex{\{'anchor': 'Swiss bank UBS announced it would cut about 1,600 more jobs at its investment bank after it posted a 8.1 billion Swiss franc loss in the fourth quarter, missing forecasts.', 'positive': "According to UBS's announcement, the bank will likely specify cutting around 1,600 more jobs at its investment bank."\}} is technically correct, but half of the main content of the anchor is omitted in the positive sentence.
In another instance, the pair does not maintain the contradiction relationship, such as   \ex{\{'anchor': 'The Red Cross reported that 400 were dead , but this was disputed by Mexican officials .', 'negative': '400 were not dead.'\}}.
The final dataset consists of 31K anchors, each with 4 negation and 2 hedging generated outputs.
We construct triples for contrastive learning by treating anchor--negation pairs as negative instances and anchor--hedging pairs as positive instances, resulting in 248K samples. We name this dataset HedgeTriple, and have made it publicly available at \url{https://hf.co/datasets/joey234/hedge_triple}.

\subsection{Contrastive triple finetuning}

Large-scale contrastive finetuning  has been shown to be an effective strategy for improving general text representations \citep{reimers-2019-sentence-bert,wang2022text}.
The key idea works by minimizing the distance between an anchor and  positive samples, and maximizing the distance between an anchor and  negative samples.
We adopt the commonly-used Multiple Negative Ranking Loss (MNRL) \citep{henderson2017efficient}, which  contrasts a positive sample against multiple negative samples.
In its original form, MNRL only requires anchor--positive pairs and randomly samples positives from other instances which are considered as negatives.
In our case, the negatives are generated explicitly, as defined above, to represent linguistic negation.
The loss function is as follows:
\begin{equation}
\mathcal{L} = -\sum_{q \in \mathcal{D}} \log\left(\frac{e^{\text{sim}(q, p^+)}}{e^{\text{sim}(q, p^+)} + \sum e^{\text{sim}(q, p^-)}}\right)
\label{eq:mnrl}
\end{equation}
where
$q$ is the query or anchor drawn from dataset $\mathcal{D}$, $p^+$ and $p^-$ are the positive and negative sample corresponding to $q$, $\text{sim()}$ is a similarity function (cosine similarity between \texttt{CLS} embeddings).

To help the model learn to distinguish between closely-related but different text, we explicitly provide hard negative samples which have high lexical overlap but contradictory meaning.
As the aim of this paper is to demonstrate the applicability of the generated triples, we did not extensively explore other contrastive learning methods but hypothesize that other contrastive losses would also work well.

\section{Experiments}

\subsection{Baseline}

\paragraph{Base model} We evaluate several leading general text encoders, namely: Sentence Transformer \citep{reimers-2019-sentence-bert} and all-mpnet-base-v2\footnote{\url{https://hf.co/sentence-transformers/all-mpnet-base-v2}} (hereafter, \textbf{MPNet}).
Our model is based on MPNet, which is a BERT-based model pretrained with masked and permuted language modeling objectives, which was further finetuned on 1B sentences pairs for embedding tasks (NLI, text similarity).

\paragraph{Negation-aware model} We evaluate two negation-aware encoder models: (1) \textbf{Jina},\footnote{\url{https://hf.co/jinaai/jina-embedding-l-en-v1}} a T5-based model finetuned on 50K triples focusing on negation;\footnote{This is the same dataset we use for selecting anchors.}
and (2) \textbf{NegMPNet},\footnote{\url{https://hf.co/tum-nlp/NegMPNet}} which is the all-mpnet-base-v2 model further finetuned on 80K pairs of sentences curated from different negation-focused datasets.

\paragraph{Our model}
We also base our method on the all-mpnet-base-v2 model, which allows for a direct comparsion. 
Our method works by finetuning the MPNet model using the contrastive loss from Eq~(\ref{eq:mnrl}) applied to our HedgeTriple dataset (see \S\ref{sec:hedge-triple-dataset}), and name the resulting model \textbf{HedgeMPNet}.

\begin{table*}[!t]
\footnotesize
    \centering
    \begin{tabular}{r  cccc}
    \toprule
         & MPNet & Jina & NegMPNet & HedgeMPNet \\
        \midrule
       \multicolumn{1}{l}{\textit{Negation benchmark}}  &  \\
        NevIR  & 8.10 & 14.61 & \underline{18.08} & \textbf{40.56} \\
        ExcluIR  & \underline{69.29} & 57.36 & 46.76 & \textbf{73.09} \\
        Cannot  & 34.91 & 30.62 & \textbf{69.44} & \underline{55.68} \\
        M3-Counterfactual & 16.20 & 41.91 & \textbf{51.29} & \underline{47.34} \\
        \midrule
        Average & 32.13  & 36.13 & \underline{46.39} & \textbf{54.17} \\ 
        \midrule
        \multicolumn{1}{l}{\textit{General benchmark}} & \\
        MTEB-Classification  & 65.07 & 67.76 & \textbf{70.83} & \underline{69.74}\\
        MTEB-PairClassification  & \underline{83.04} & \textbf{84.80} & 79.05 & 82.20\\
        MTEB-Reranking  & \textbf{68.83} & 56.42 & \underline{68.24} & 66.85\\
        MTEB-Clustering  & \textbf{43.69} & 37.15  & \underline{38.45} & 36.88\\
        MTEB-Retrieval  & \underline{43.10} & \textbf{44.81} & 36.12 & 35.75\\
        MTEB-STS  & \underline{80.28} & \textbf{80.96}  & 77.58 & 77.49\\
        MTEB-Summarization & 27.49 & \underline{29.58} & 27.49 & \textbf{30.98} \\
        \midrule
        Average (56 datasets) & \textbf{58.79} & \underline{57.38} & 56.82 & 57.14 \\
        
    \bottomrule
    \end{tabular}
    \caption{Results on negation and general benchmarks. The reported score for each task is the main metric to evaluate that task; higher is better. \textbf{bold} and \underline{underline} denotes the best and second-best scores respectively.}
    \label{tab:main_results}
\end{table*}

\subsection{Benchmarks}


\subsubsection{Negation-focused benchmarks}

\paragraph{NevIR \citep{weller-etal-2024-nevir}:} an information retrieval benchmark, based on CONDAQA \citep{ravichander-etal-2022-condaqa}. Each sample consists of a pair of contrasting queries, each with one relevant document.
The goal is to correctly rank the two documents with respect to each query.
We report the Right Rank (RR) metric, which is the percentage of time the models correctly produce the correct rank for the pair of queries, with chance performance of 25\%, as for each data sample, the model needs to correctly rank 2 queries.

\paragraph{ExcluIR \citep{zhang2024excluirexclusionaryneuralinformation}:} a benchmark focusing on exclusion queries (e.g.\ \ex{Apart from Old \& Kumar Go to White Castle, what other films has actor Errol Sitahal appeared in?}). The dataset is a modified version of HotpotQA \citep{yang-etal-2018-hotpotqa}.
We also use RR here, with chance performance of 50\% as each query is separately evaluated.

\paragraph{Cannot \citep{anschutz-etal-2023-correct}:} an MT evaluation dataset, where negation is a common cause of error. The dataset includes sentence pairs and their semantic similarity scores. We report Spearman's correlation $\rho$ between our model predictions (cosine similarity) and the ground truth.


\paragraph{M3-Counterfactual \citep{otmakhova-etal-2022-m3}:}  a subset of the M3 dataset, constructed by manually corrupting statements in biomedical literature to evaluate model's robustness in a counterfactual setting. The modification includes adding negation to statements, changing statements into non-evidential sentences (\ex{There is no evidence that ...}), or changing the modality (e.g.\ by adding hedging words such as \ex{might} or intensifiers such as \ex{certainly}). We reformat the data into text-similarity-style task and assign original--negation pairs a score of $-1$, original--no evidence pairs a score of $0$, and original--hedged pairs a score of $1$.
Similar to the Cannot dataset, we evaluate the models' performance using Spearman's correlation $\rho$ against the cosine similarity estimates.

\subsubsection{General benchmarks}

Aside from negation benchmarks, we also evaluate the general capabilities of finetuned models on standard English benchmarks.
Specifically, we use the 
comprehensive general benchmark set of text understanding tasks \textbf{MTEB} \citep{muennighoff2022mteb}, spanning 7 subtasks with 56 datasets. 

\section{Main findings}

\subsection{Negation and general benchmark results}

As can be seen in \Cref{tab:main_results}, in general our model (``HedgeMPNet'') outperforms all similar-sized text embedding models on negation benchmarks, while maintaining similar performance on general benchmarks.
On the negation side, we see large increases on both NevIR and ExcluIR over both general (all-mpnet-base-v2) and negation-focused models (Jina and NegMPNet).
Note that the high performance of NegMPNet on Cannot is because it is in-domain data for this model, in that the model was fine-tuned on the training portion of the same dataset.
Over general benchmarks, we can observe increases on classification and summarization tasks, and drops on other tasks.
One interesting pattern is the large increase on sentiment classification datasets inside MTEB-Classification, showing that this strategy is especially helpful for sentiment-related tasks where people tend to express opinions subjectively (using more hedging) and using terms associated with negation to express negative sentiment.

We further conducted additional experiments to ablate the impact of the HedgeTriple dataset. To save time and resources, for subsequent ablation experiments, we only evaluate on a subset of MTEB that has been shown to correlate highly with overall model performance, as introduced in \citet{llm2vec}.

\begin{table*}[!t]
\footnotesize
    \centering
    \begin{tabular}{r  cccc}
    \toprule
         & MPNet & Only negation & Only hedging & HedgeMPNet \\
        \midrule
       \multicolumn{1}{l}{\textit{Negation benchmark}}  &  \\
        NevIR  & 8.10 & \underline{35.72} & 9.83 & \textbf{40.56} \\
        ExcluIR  & 69.29 & \textbf{76.10} & 64.83 & \underline{73.09} \\
        Cannot  & 34.91 & \underline{54.89} & 15.28 & \textbf{55.68} \\
        M3-Counterfactual & 16.20 & \textbf{53.17} & 31.31 & \underline{47.34} \\
        \midrule
        Average & 32.13 & \textbf{55.57} & 29.98 & \underline{54.17} \\ 
        \midrule
        \multicolumn{1}{l}{\textit{General benchmark}} & \\
        MTEB-Classification-lite  & 64.56 & \underline{65.24} & 63.9 & \textbf{66.30} \\
        MTEB-PairClassification  & \underline{90.15} & 85.17 & \textbf{94.58} & 88.92 \\
        MTEB-Reranking  & \textbf{70.32} & 67.65 & \underline{69.11} & 68.25 \\
        MTEB-Clustering & \textbf{39.27} & 36.22 & \underline{36.90} & 32.79 \\
        MTEB-Retrieval & \underline{48.46} & 30.88 & \textbf{48.79} & 34.61 \\
        MTEB-STS & \textbf{84.87} & 78.36 & \underline{83.58} & 79.07 \\
        MTEB-Summarization & 27.49 & \underline{30.77} & 30.17 & \textbf{30.98} \\
        \midrule
        Average (16 datasets) & \underline{60.73} & 56.33 & \textbf{61.00} & 57.27 \\
        
    \bottomrule
    \end{tabular}
    \caption{Ablation results on negation and general benchmarks. The reported score for each task is the main metric to evaluate that task; higher is better. "Only negation" and "Only hedging" refer to the setting of finetuning MPNet on only negation data and hedging data, respectively.}
    \label{tab:hedging_vs_negation}
\end{table*}

\paragraph{Balancing negation--general capability tradeoffs}

\begin{figure}[!t]
    \centering
    \includegraphics[width=\linewidth]{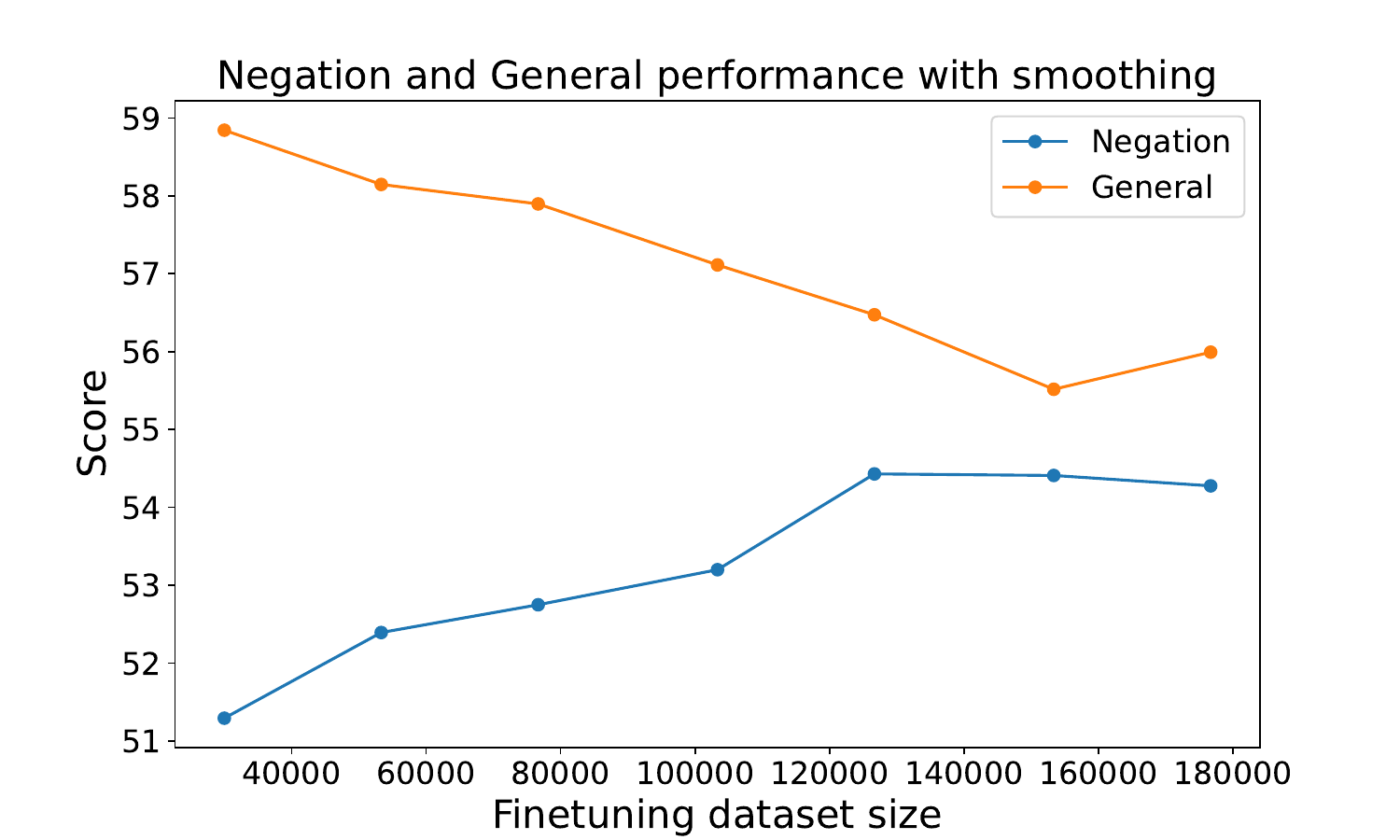}
    \caption{Negation--general performance tradeoff when finetuned on different dataset sizes, smoothed with moving average with window size 3} %
    \label{fig:tradeoff}
\end{figure}

Catastrophic forgetting, where a model loses some of its original capabilities, is inevitable when models are further finetuned to adapt to new tasks or domains.
Thus, we experiment with finetuning using different data sizes ranging from 10K to 200K instances, to observe the impact of training data size. 
Results show that finetuning on more HedgeTriple samples leads to larger performance gains on negation benchmarks at the cost of general capabilities. 
From \Cref{fig:tradeoff}, we can see that performance on negation benchmarks is observed with as few as 10K samples on HedgeTriple.
The optimal point to balance out the tradeoff is around 150K training samples.
We hypothesize that retention of general capabilities is thanks to 
exposure 
to hedging data, and conduct an ablation analysis with respect to data attribution, i.e.\ finetuning only with either hedging or negation data, to further investigate this.

\paragraph{Data attribution}

\begin{figure}[!t]
    \centering
    \begin{minipage}{0.45\textwidth}
        \centering
        \includegraphics[width=\textwidth]{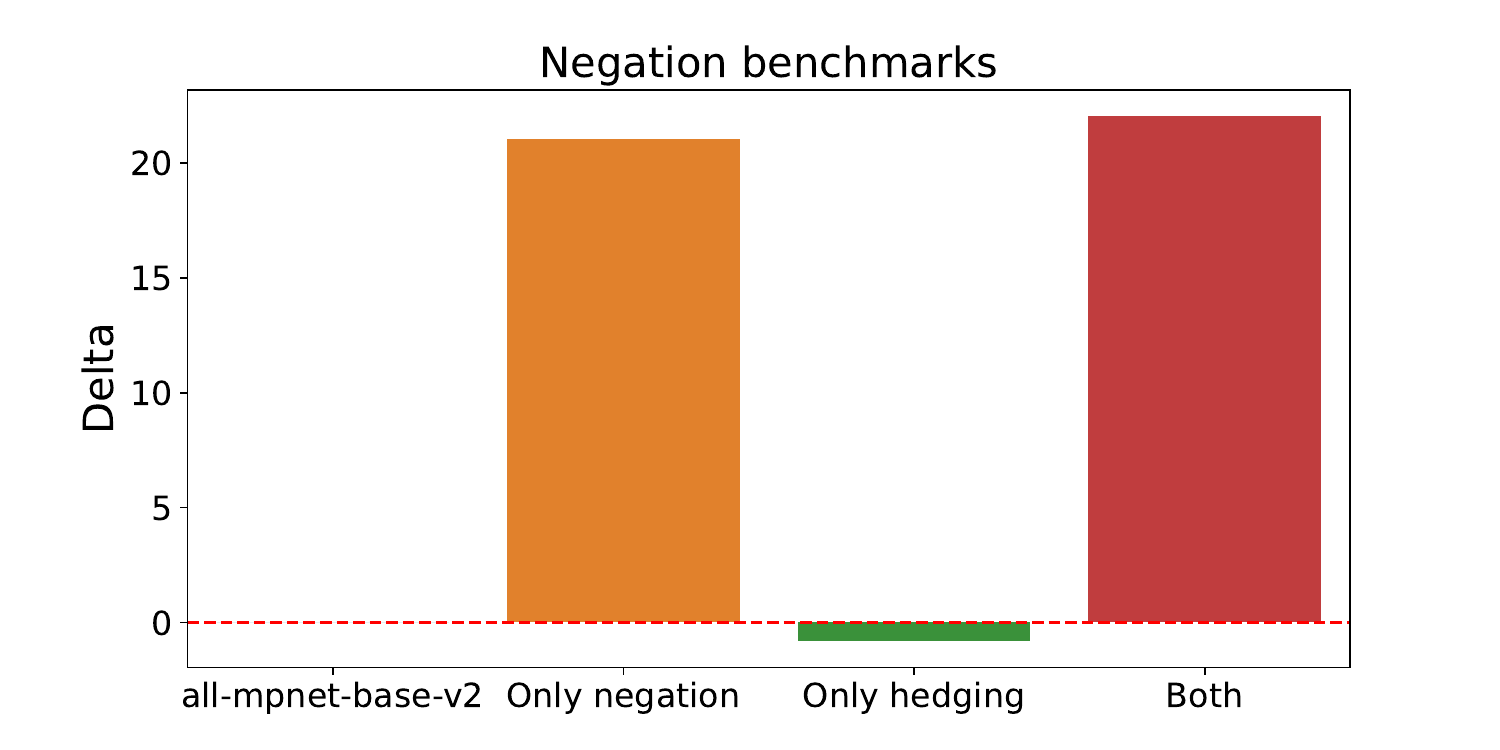}
        \label{fig:ablate_negation}
    \end{minipage}
    \hfill
    \begin{minipage}{0.45\textwidth}
        \centering
        \includegraphics[width=\textwidth]{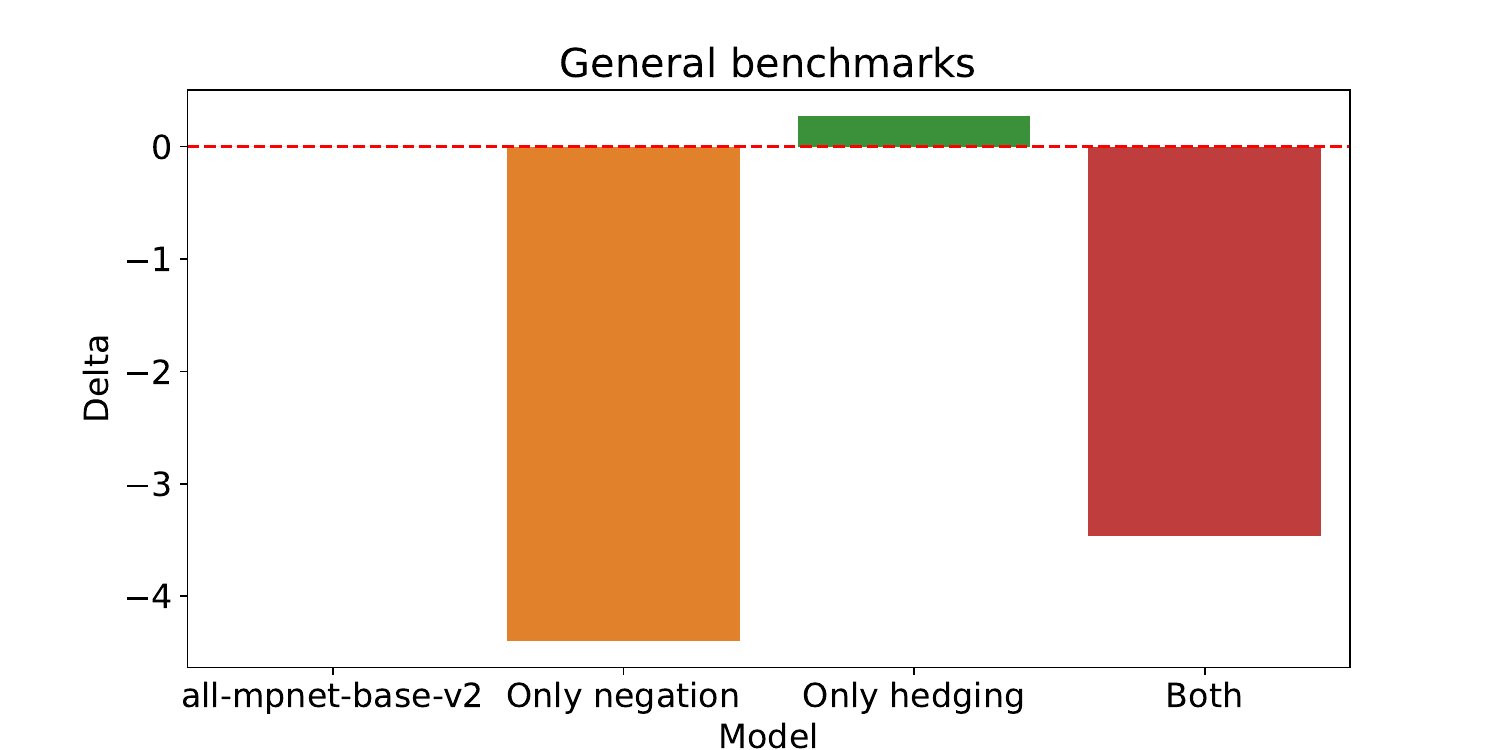}
        \label{fig:ablate_general}
    \end{minipage}
    \caption{Relative difference wrt. all-mpnet-base-v2 using different portions of data}
    \label{fig:ablation}
\end{figure}

        

We conducted an ablation study to evaluate the impact of each portion of the data: only using negation data (``Only negation''), only using hedged data (``Only hedging''), or both (\Cref{tab:hedging_vs_negation}).
When only using negation data, we used the original positive sentences from the negation dataset which we sampled the anchors from; while for only hedged data, we used the original negative sentences.
The results show (\Cref{fig:ablation}) that combining both data types leads to the best performance, and that negation data plays a more important role.
Only using hedging data is not beneficial as all the benchmarks considered are more focused on negation, and do not have any explicit measure for hedging.
However, finetuning on hedging data is beneficial in retaining general capabilities, with high results on the MTEB-lite set, surpassing even the base model without additional finetuning.

\paragraph{Data contamination}

We found no exact matches between any of the negation benchmarks and HedgeTriple.\footnote{Defined as when a sample has a text field (query, doc, text, etc.) that is included in HedgeTriple, or vice versa.}
$N$-gram analysis reveals that overlap happens for less than 4\% of the samples in all datasets (noting all samples have a minimum length of 5 words).
Moreover, the overlap here happens in the retrieval corpus, not on the query set. This is standard in IR and is not considered data contamination.
Hence data leakage is negligible. 


\begin{figure}[!t]
    \centering
    \begin{minipage}{0.45\textwidth}
        \centering
        \includegraphics[width=\textwidth]{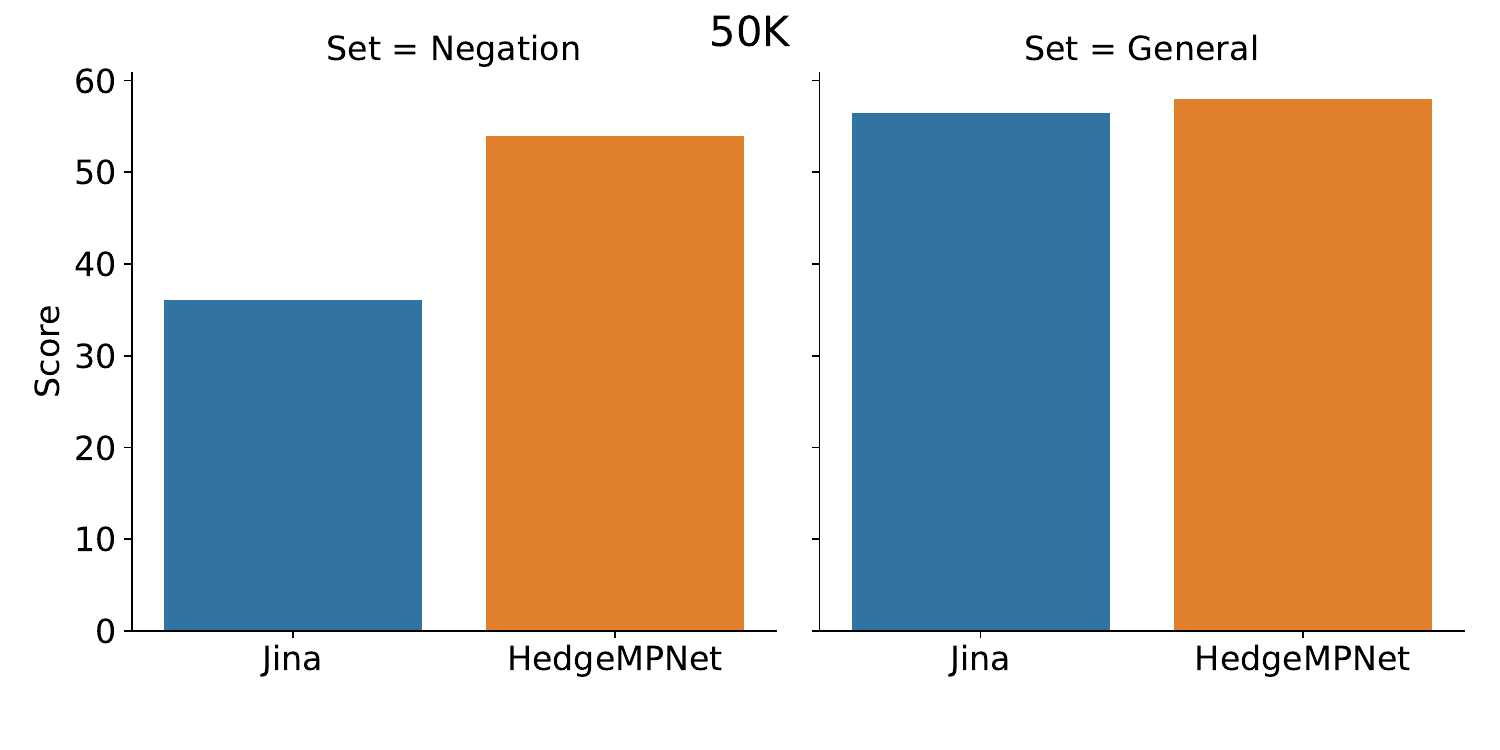}
        \label{fig:diversity_50k}
    \end{minipage}
    \hfill
    \begin{minipage}{0.45\textwidth}
        \centering
        \includegraphics[width=\textwidth]{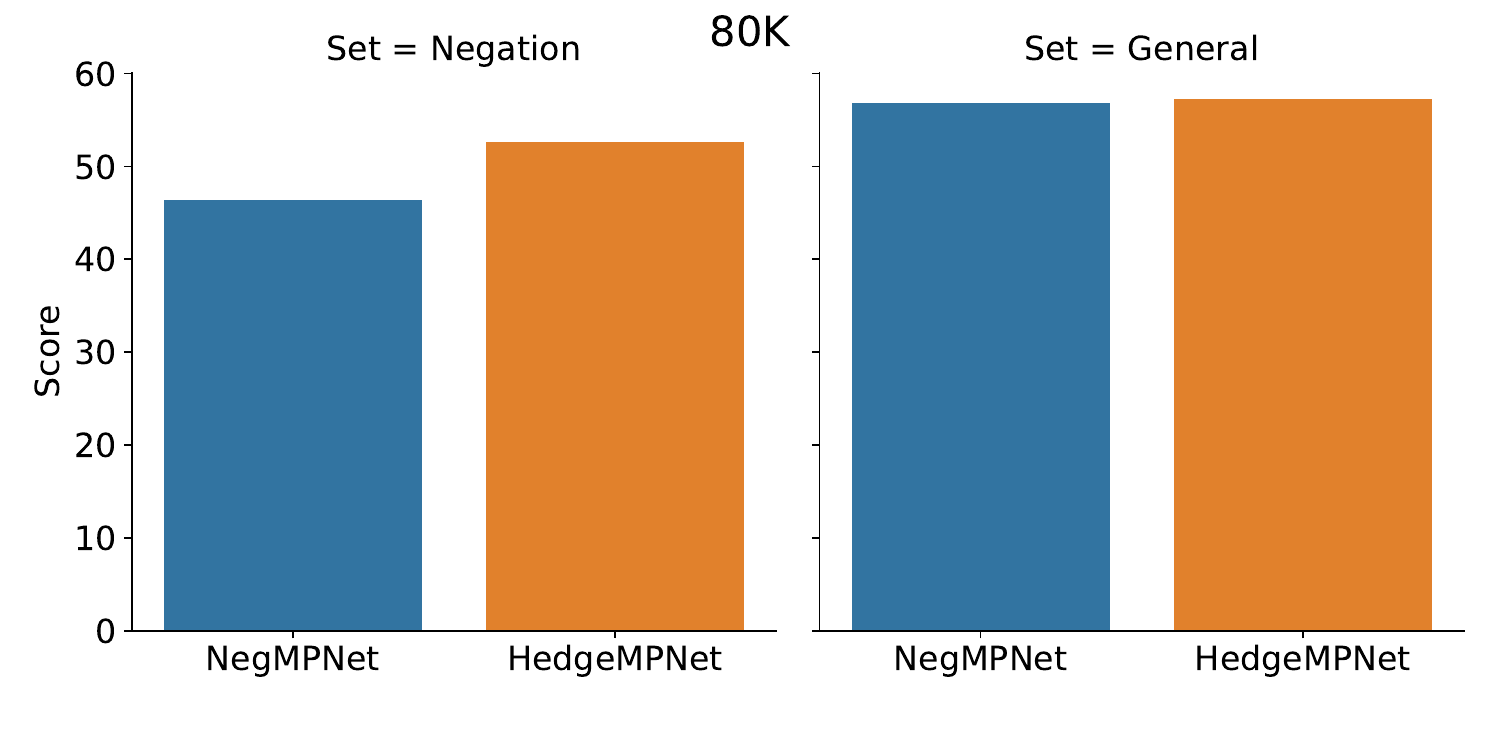}
        \label{fig:diversity_80k}
    \end{minipage}
    \caption{HedgeMPNet compared with similar negation-focused models finetuned on similar-sized datasets.}
\end{figure}

\paragraph{Diversity vs.\ quantity}

As detailed in \Cref{sec:related-work}, both negation and finetuning using contrastive learning to improve SBERT have been explored in previous work.
However, they only consider the most straightforward types of negation: syntactically adding \ex{not} to the main verb either by rules in CANNOT \citep{anschutz-etal-2023-correct}, or using GPT-3.5 in Jina Embedding \citep{günther2023jina}. 
Instead, a main contribution of this work is the adoption of linguistically-sound taxonomies to create more diverse negation data.
Our models finetuned on similar data sizes outperform both NegMPNet ($\sim$80K samples) and JinaAI Embedding model ($\sim$50K samples) on the negation benchmarks.
This shows that diversity in negation and hedging patterns plays a bigger role than quantity.

\begin{table*}[!t]
\footnotesize
    \centering
    \begin{tabular}{r ccccc}
    \toprule
         & HedgeMPNet & Llama-3-8B-Instruct & Only negation & Only hedging   &  Llama-3-8B-Hedge \\
        \midrule
       \multicolumn{1}{l}{\emph{Negation benchmark}}  &  \\
        NevIR (0shot) & 40.56 &  \underline{74.04}  & 73.13 & 68.13 &  \textbf{78.13}  \\
        ExcluIR (0shot)  & 73.09 & 91.71  & \textbf{93.83} & 92.50 & \underline{93.40} \\
        CANNOT (0shot)  & 55.68 & 44.16  & \textbf{63.99} & 53.22 & \underline{60.89} \\
        M3-Counterfactual & 47.34 &  56.05  & 68.59 & \underline{69.64} & \textbf{76.27} \\
        \midrule
        Average & 54.17 & 66.85 & \underline{75.15}  & 70.61  & \textbf{77.17}  \\ 
        \midrule
        \multicolumn{1}{l}{\emph{General benchmark}}  & \\
        MMLU (5shot) & N/A &  \textbf{65.68} & 63.59 & \underline{63.65} & 63.03 \\
        HellaSwag (0shot) & N/A & \underline{75.77} & \textbf{77.17} & 76.80 & 75.31 \\
        GSM8K (5shot, CoT) & N/A & \textbf{75.36} & 66.41 & \underline{70.96} & 67.10 \\
        \midrule
        Average & N/A & \textbf{72.27} & 69.06 & \underline{70.47} & 68.48 \\
        
    \bottomrule
    \end{tabular}
    \caption{LLM results on negation and general benchmarks in comparison with the best performing model from the previous experiment. The reported score for each task is the task-specific main metric 
    "Only negation" and "Only hedging" refer to the setting of finetuning Llama-3-8B-Instruct on only negation data and hedging data, respectively.}
    \label{tab:llm_results_ablate}
\end{table*}

\subsection{Effect of HedgeTriple on LLMs}

We also look at the performance of a current-gen decoder-only LLM (Llama-3-8B-instruct) on negation benchmarks, and whether finetuning it on HedgeTriple can improve its handling of negation and hedging (\Cref{tab:llm_results_ablate}).
We treat the task as ranking between two documents, with the following prompt:

\begin{prompt}
Document 1: \textbf{doc 1}\\
Document 2: \textbf{doc 2}\\
Query: \textbf{q}\\

Which document is more relevant to the query? Please choose 1 or 2.\\
Answer: 
\end{prompt}

For the  CANNOT similarity task, we ask the model to score the sentence pairs:

\begin{prompt}
Determine the similarity between the following two sentences (S1, S2).
The score should be ranging from -1.0 to 1.0, and can be a decimal. \\
S1: \textbf{sentence 1}\\
S2: \textbf{sentence 2}\\
Score: 
\end{prompt}

Simply applying the LLM in a zero-shot manner, we immediately see much higher performance than HedgeMPNet on both NevIR and ExcluIR.
However, Llama-3-8B-instruct is 
several orders of magnitude larger in parameter size, and much more expensive to apply as a text encoder.
Regardless, there is active research on deriving text embeddings from LLMs, such as via bidirectional text encoders \citep{llm2vec,wang2024improvingtextembeddingslarge}.

Next, we convert HedgeTriple into pairs to finetune Llama-3-8B-Instruct with LoRA \citep{hu2022lora} (finetuning details in \Cref{sec:llm_finetune_details}).
For instance, a triple is converted into two pairs:

\begin{prompt}
Sentence 1: \textbf{A boy holding his skateboard behind him and covering his behind.}\\
Sentence 2: \textbf{The boy is sitting comfortably without his skateboard and with his behind exposed.}\\
Do the two sentences have opposite meaning? Yes or No.\\
Answer: \textbf{Yes}\\

Sentence 1: \textbf{A boy holding his skateboard behind him and covering his behind.}\\
Sentence 2: \textbf{The boy, it seems, held his skateboard behind him and covered his behind.}\\
Do the two sentences have opposite meaning? Yes or No.\\
Answer: \textbf{No}
\end{prompt}

We observe further improvements in the finetuned model (Llama3-8B-Hedge) over the base version, showing that the HedgeTriple is also beneficial for current-gen LLMs.
Despite there still being room for improvement, overall, LLMs appear to be able to distinguish between negated and non-negated contexts quite well when evaluated in a pairwise setting.
However, this finding may not generalize to other negation benchmarks, which LLMs still struggle with \citep{truong-etal-2023-language}.

In addition, we evaluate the general capabilities of the fine-tuned LLM on three common benchmarks --- MMLU \citep{hendryckstest2021}, HellaSwag \citep{zellers-etal-2019-hellaswag}, and GSM8K \citep{cobbe2021gsm8k} --- to determine if the fine-tuning has led to catastrophic forgetting.
We use the default settings for each benchmark in lm-evaluation-harness \citep{eval-harness}.
Overall, we observe comparable performance with and without fine-tuning for MMLU and HellaSwag, but a drop on GSM8K (which contains grade school math problems).
We also notice a degradation over the MMLU subset related to mathematics (e.g.\ high school/college/elementary mathematics, statistics).
This finding implies that that negation robustness can negatively impact the arithmetic reasoning abilities of models.
We conducted an error analysis on GSM8K and found that most of the errors are due to wrong calculations---even though the equations are correct---and the loss of quantitative commonsense knowledge (see \Cref{sec:gsm8k_error} for details).

We also performed ablation to see the impact of negation and hedging data on the LLM's capabilities (Only negation and Only hedging in \Cref{tab:llm_results_ablate}). 
Similar to the encoder models experiments, we notice a larger effect of negation data in improving negation understanding capabilities, but combining both negation and hedging leads to the best scores.
Over the general benchmarks, hedging data also leads to best retention of general model capabilities.
Interestingly, combining both data types leads to worse results compared with using either alone.

\section{Conclusion}
Negation and hedging are important phenomena that have huge impact on language understanding but are often overlooked when evaluating models' capabilities.
In this work, we propose a strategy to improve text embedding robustness to negation and hedging based on contrastive finetuning on synthetic data distilled from LLM.
Our prompts are carefully crafted with well-defined linguistic taxonomies to ensure diversity in the negation and hedging patterns.
We conducted extensive experiments and observed drastic improvements on negation benchmarks while retaining general capabilities.
Furthermore, finetuning an LLM on the generated triples is also beneficial in improving negation understanding abilities, at the cost of a small degradation in mathematical performance.

\clearpage

\section{Limitations}

\paragraph{Prompting}
For data generation, we iteratively update the prompts based on manually inspecting the output of LLMs until observing the desired behaviour.
Employing automatic prompt optimization technique suchs as DSPy \citep{khattab2023dspy} would result in better prompts but we decided not to explore this as the current results are satisfactory.

\paragraph{Other languages}
As a starting point, we focused exclusively on English, but the same strategy can be readily adapted to other languages. Thus, we claim that the findings of this work are generalizable to a multilingual setting.

\paragraph{Finetuning strategies}
In both contrastive finetuning of text encoders and LLM supervised finetuning, we experimented with a relatively simple and straightforward strategy and data format. For the LLM, finetuning using more diverse instructions with a reasoning step would likely unlock more sophisticated negation reasoning abilities.

\bibliography{custom}

\appendix
\section{List of hedge cues}
\label{sec:hedge_cues}

\paragraph{Single-word cues}
\textsf{['wish', 'conjecture', 'wonder', 'implying', 'unlikely', 'likely', 'slight', 'likelihood', 'possibly', 'sufficient', 'question', 'whether', 'believe', 'wouldnt', 'expect', 'hinting', 'hope', 'suspect', 'if', 'afraid', 'necessarily', 'thinking', 'expecting', 'might', 'apparent', 'felt', 'apparently', 'seem', 'may', 'certainly', 'propose', 'probable', 'imply', 'potentially', 'shouldnt', 'nearly', 'suggestive', 'impression', 'clear', 'can', 'or', 'hesitant', 'probability', 'specify', 'hopefully', 'clean', 'sure', 'ought', 'wrong', 'why/if', 'argue', 'somewhat', 'unsure', 'plausible', 'doubtful', 'must', 'anticipate', 'uncertainty', 'feel', 'clearly', 'either', 'specifying', 'appreciate', 'appear', 'indication', 'couldnt', 'hoping', 'possibility', 'cant', 'suggesting', 'proposing', 'notion', 'presumably', 'potential', 'seemingly', 'doubt', 'uncertain', 'probably', 'assume', 'undoubtedly', 'assumption', 'sense', 'surely', 'arguing', 'cannot', 'clearer', 'should', 'debatable', 'indicating', 'indicate', 'strange', 'speculate', 'weird', 'suggestion', 'think', 'suppose', 'arguably', 'questionable', 'would', 'imagine', 'claim', 'theoretically', 'maybe', 'suggest', 'presume', 'idea', 'like', 'unclear', 'implication', 'almost', 'unknown', 'possible', 'appearence', 'rather', 'implicit', 'puzzling', 'supposedly', 'suspicion', 'impossible', 'wondering', 'argument', 'vague', 'thought', 'hypothesize', 'seeming', 'could', 'guessing', 'tend', 'say', 'wether', 'maynot', 'slightly', 'feeling', 'assuming']}

\paragraph{Multi-word cues}
\textsf{['not very clear', 'not surely', 'cannot claim', 'seeming like', 'not clear', 'on the fence', 'not so sure', 'not very sure', 'hard to pin down exactly', 'look like', 'felt like', 'not also sure', 'not really sure', 'not totally sure', 'cannot imagine', 'isnt clear', 'not completely sure', 'not exactly sure', 'no idea', 'not entirely clear', 'could not figure out', 'not at all sure', 'wonder if', 'do not convincingly', 'mostly clear', 'feel like', 'cannot hope', 'not 100 \% sure', 'sound like', 'not clearly', 'not convincing', 'not at all clear', 'not conclusive', 'not quite sure', 'not entirely sure', 'can not', 'not totally clear', 'not all are clear', 'somewhat unclear', 'not even sure', 'very unclear', 'seem like', 'can imagine', 'not certain', 'not sure']}

\section{LLM finetuning details}

\label{sec:llm_finetune_details}

\begin{table}[!htbp]
    \centering
    \begin{tabular}{p{2cm}|p{4cm}}
    \toprule
        Parameter & Value \\
        \midrule
        Epoch & 20 \\
        Rank & 16 \\
        Target modules & q\_proj, k\_proj, v\_proj, o\_proj, gate\_proj, up\_proj, down\_proj \\
        
        Alpha & 16 \\
        Dropout &  0 \\
        Bias &  None \\
        \bottomrule
    \end{tabular}
    \caption{Caption}
    \label{tab:finetuning-details}
\end{table}

\section{A closer look at the GSM8K output}
\label{sec:gsm8k_error}

We compare the CoT output of Llama-3-8B-Instruct and our finetuned version, Llama-3-8B-Hedge, to identify the reasons for the drop in arithmetic reasoning skills.
We randomly sample a set of 100 samples where Llama-3-8B-Hedge get the answer wrong, and classify the error types into different categories and summarize the results in \Cref{tab:gsm8k_error}. From this analysis, we can see that the finetuned model regresses somewhat in calculation and commonsense knowledge. Another large number of cases come from a bug in the evaluation toolkit rather than actual errors of the model. One interesting pattern is that model tends to do some inverted calculations, messing up the correct position of the two operands, affecting subtraction and division calculations.  Although not very prevalent, we also notice a small portion of cases where the base model gets the correct answer despite incorrect reasoning steps.

\begin{table*}[]
\footnotesize
    \centering
    \begin{tabular}{p{2cm}p{3cm}p{5cm}p{5cm}p{0.5cm}}
    \toprule
        \textbf{Error type}  & \textbf{Description} & \textbf{Llama-3-8B-Instruct} &  \textbf{Llama-3-8B-Hedge} &  \textbf{\%}  \\
        \midrule

        Wrong calculation & When the reasoning step and the final equation to get the result are correct, but the calculated results are wrong & ... The total cost of the trip is \textbf{\$5000 + \$5000 + \$5400 = \$15400} ... & ...
The total cost of the trip is \textbf{\$5400 + \$10000 = \$14000}.
...
& 20 \\
\midrule
        Fail to include all information & When the logic of the solution is correct but miss one (or more) important steps, leading to incorrect final results & The doll cost as much as 3 action figures, so the doll costs 3 x \$5 = \$15.
One red car costs \$4, so 5 red cars cost 5 x \$4 = \$20.
The total cost of all toys is \textbf{\$15 + \$20 + 3 x \$5 = \$15 + \$20 + \$15 = \$50}. & The doll cost as much as 3 action figures, so the doll costs 3 x \$5 = \$15.
One red car costs \$4, so the 5 red cars cost 5 x \$4 = \$20.
The total cost of the toys is \textbf{\$15 + \$20 = \$35} & 20\\
\midrule
        Wrong unit & Error in parsing the final answers  & Violetta needs 5 crayons and each crayon costs \$2, so she needs 5*2 = <<5*2=10>>10 dollars.
She has \$20 and will spend \$10, so she will get \$20-\$10 = <<20-10=10>>\textbf{10} dollars in change & Violetta needs 5 crayons, and each crayon costs \$2. So she needs 5 * 2 = 10 dollars.
She has \$20, so she will get \$20 - \$10 = \textbf{\$10} in change & 19 \\
\midrule
        Fail to reason with multiplication & When the reasoning fails to take into account the quantitative commonsense knowledge (e.g. a week has 7 days, a cow has 4 legs) & Melissa has 8 + 5 + 3 + 12 = <<8+5+3+12=28>>28 animals to groom over the next week.
\textbf{Therefore, She will groom 28 / 7 = <<28/7=4>>4 animals each day.} & She has 8 dogs, 5 cats, 3 birds, and 12 horses that need to be groomed. In total, she has 8 + 5 + 3 + 12 = 28 animals that need to be groomed.
\textbf{She will split the grooming jobs evenly over the days, so she will groom 28 animals each day of the week} & 18 \\
\midrule
        Invert calculation & When the model perform subtraction rather addition or switch positions of the left and right operands & On Thursday, the mechanic earned 6 * $60 = $360 for truck tires and 4 * \$40 = \$160 for car tires.  So the total revenue on Thursday was \$360 + \$160 = \$520.
On Friday, the mechanic earned 12 * \$40 = \$480 for car tires.  So the total revenue on Friday was \$480.
The mechanic earned \$520 on Thursday and \$480 on Friday.  \textbf{The difference is \$520 - \$480 = \$40}. & On Thursday, the mechanic earned 6 * \$60 = \$360 for truck tires and 4 * \$40 = \$160 for car tires.  So, the total revenue on Thursday is \$360 + \$160 = \$520.

On Friday, the mechanic earned 12 * \$40 = \$480 for car tires.  So, the total revenue on Friday is \$480.

\textbf{The mechanic earned \$480 - \$520 = -\$40} more on the day with lower revenue. & 10 \\
\midrule
        Hallucination & When the reasoning chain is not coherent, logical or relevant to the question & Christina had 8 good days, 8 bad days, and 8 neutral days. This is a total of 8 + 8 + 8 = <<8+8+8=24>>24 days.
She also had 3 more days, which are 1 good, 1 neutral, and 1 good. This is a total of 24 + 3 = <<24+3=27>>27 days.
She had 30 days in total, so she has 30 - 27 = <<30-27=3>>3 days left.
Of these 3 days, 2 are good, so she has 2 good days left. & \textbf{Christina had 12 good days in total.
The first 8 days were good, so that's 8 good days.
The next 3 days were good, so that's 3 good days.}
12 - 8 - 3 = 1 good day left in the month. & 5 \\
\midrule
        Base model is wrong & When ground truth is incorrect, or when base model gives correct answers but the reasoning chain is wrong,  & The interest for the first month is \textbf{2\% of \$100 = <<2\%*100=2>>\$2}
The interest for the second month is \textbf{2\% of \$100 = <<2\%*100=2>>\$2}
The interest for the third month is \textbf{2\% of \$100 = <<2\%*100=2>>\$2}
The total interest is \textbf{\$2 + \$2 + \$2 = <<\$2+\$2+\$2=6>>\$6}
The total amount Mandy should pay is \textbf{\$100 + \$6 = <<\$100+\$6=106>>\$106} & After 1 month, Mandy owes Benedict \textbf{\$100 + 2\% of \$100 = \$100 + \$2 = \$102}.

After 2 months, Mandy owes Benedict \textbf{\$102 + 2\% of \$102 = \$102 + \$2.04 = \$104.04}.

After 3 months, Mandy owes Benedict \textbf{\$104.04 + 2\% of \$104.04 = \$104.04 + \$2.0816 = \$106.1216}.

Mandy should give Benedict \$106.1216.

  & 4 \\
\midrule
        Repetition & When the model repeats one reasoning step many times. & Let the cost of the lot be x. Then the cost of the house is 3x.
The total cost is 120,000, so x + 3x = 120,000.
Combine like terms: 4x = 120,000.
Divide by 4: x = 30,000.
The cost of the house is 3x = 3 * 30,000 = \$90,000  & The house cost three times as much as the lot, so the house cost 3 x (the cost of the lot) = 3 x (the cost of the house + the cost of the lot - the cost of the house) = 3 x (the cost of the lot) = 3 x (the cost of the house + the cost of the lot - the cost of the house) = 3 x (the cost of the lot) ... & 4 \\
    \bottomrule
    \end{tabular}
    \caption{Error analysis of the sample of 100 errors made by Llama-3-8B-Hedge on GSM8K.}
    \label{tab:gsm8k_error}
\end{table*}






\end{document}